\documentclass{article}
\usepackage{spconf}
\usepackage{graphicx}
\usepackage{amsmath}
\usepackage{amssymb}
\usepackage{booktabs}
\usepackage{csquotes}
\usepackage{microtype}
\usepackage{nicefrac}

\usepackage[ruled,vlined,linesnumbered]{algorithm2e}

\SetCommentSty{mycommfont}

\usepackage[breaklinks,colorlinks]{hyperref}

\usepackage{color}

\title{AdaWCT: Adaptive Whitening and Coloring Style Injection}
\name{Antoine Dufour\textsuperscript{\textdagger}, Yohan Poirier-Ginter\textsuperscript{\textdagger}, Alexandre Lessard\textsuperscript{\textdaggerdbl}, 
Ryan Smith\textsuperscript{\textdaggerdbl},}
\secondlinename{Michael Lockyer\textsuperscript{\textdaggerdbl}, 
and Jean-Fran\c{c}ois Lalonde\textsuperscript{\textdagger}\thanks{Contact: \texttt{jflalonde@gel.ulaval.ca}.}}
\address{\textsuperscript{\textdagger}Université Laval, \textsuperscript{\textdaggerdbl}Gearbox Studios}

\begin{document}
%\ninept

\maketitle

\begin{abstract}
Adaptive instance normalization (AdaIN) has become the standard method for style injection: by re-normalizing features through scale-and-shift operations, it has found widespread use in style transfer, image generation, and image-to-image translation. In this work, we present a generalization of AdaIN which relies on the whitening and coloring transformation (WCT) which we dub AdaWCT, that we apply for style injection in large GANs. We show, through experiments on the StarGANv2 architecture, that this generalization, albeit conceptually simple, results in significant improvements in the quality of the generated images. 
\end{abstract}

\begin{keywords}
Image-to-image translation, generative adversarial networks, whitening and coloring transformation. 
\end{keywords}

%!TEX root = main.tex
\section{Introduction}

Generative Adversarial Networks (GANs)~\cite{goodfellow2014generative} have become what is perhaps the most important family of methods for image generation and manipulation these past years. In particular, image-to-image translation (i2i) methods~\cite{Isola2017ImagetoImageTW} seek to learn a mapping between two related domains, thereby ``translating'' and image from one domain to another while preserving some information from the original. Here, one key idea is to enforce the model to preserve the ``content'' of the image while modifying its ``style''. In this context, Adaptive Instance Normalization (AdaIN)~\cite{huang2017arbitrary} has become the standard method for style injection in i2i and has been exploited in popular architectures such as StarGAN~\cite{choi2018stargan} and StyleGAN~\cite{StyleGAN2018}.
  
AdaIN works by modulating the statistics of feature maps inside the network: it first normalizes them by subtracting their mean and dividing by their standard deviation; then it injects a (learned) style vector by the reverse of this operation. Typically, the injected style vector is provided by a mapping network~\cite{StyleGAN2018}. Although this method has been used successfully in a variety of image-to-image translation scenarios, the statistical representation of AdaIN is limited in that it does not take into account the existing correlations between the feature maps of the generator. However, in the style transfer literature, the Whitening \& Coloring Transformation (WCT) has become the preferred approach precisely because of its capacity to take into account these correlations. Yet, despite its good performance in terms of style transfer, WCT has so far not been used for style injection. 

In this work, we fill this gap by replacing AdaIN with an explicit WCT operation for style injection in GANs. This change, dubbed ``Adaptive WCT'' (AdaWCT), is a generalization of AdaIN that can be used as a drop-in replacement for the AdaIN blocks (without any additional change) in existing GAN architectures and we present its impact on image-to-image translation tasks. Indeed, in conditional image generation tasks where the latent space is intended to represent the style of the images, we find that AdaWCT helps in disentangling style from content and significantly improves image quality---results that we demonstrate through quantitative and qualitative evaluations on the StarGANV2~\cite{choi2020stargan} architecture. 
%whitening helps ensure that the space encodes only stylistic information which allows the content of the input image to be more visible. 

%!TEX root = main.tex
\section{Related work}

Generative adversarial networks (GANs)~\cite{goodfellow} have demonstrated promising results in various applications in computer vision, including image generation~\cite{StyleGAN2018,karras2020analyzing,karras2021alias} and image-to-image translation~\cite{MUNIT,pix2pix,DRIT,msgan,cycleGan}. 
Recent work has striven to improve sample quality and diversity, notably through theoretical breakthroughs in terms of defining loss functions which provide more stable training~\cite{Wgan,wgangp,lsgan} and encourage diversity in the generated images~\cite{diverseGan}. Architectural innovations also played a crucial role in these advancements~\cite{choi2018stargan,ProGanCelebA}. 
%For instance, \cite{sagan} makes use of an attention layer that allows it to focus on long-range dependencies present in the image. Spectral normalization~\cite{miyato2018spectral} stabilizes the network, which also translates into having high quality samples. 

Also dubbed ``conditional''~\cite{CItoI}, reference-guided i2i translation methods seek to learn a mapping from a source to a target domain while being conditioned on a specific image instance belonging to the target domain. In this case, some methods attempt to preserve the ``content'' of the source image (identity, pose) and apply the ``style'' (hair/skin color) of the target. Inspired by the mapping network of StyleGAN~\cite{StyleGAN2018}, recent methods~\cite{choi2018stargan,choi2020stargan,MUNIT} make use of a style encoder to extract the style of a target image and feed it to the generator, typically via adaptive instance normalization (AdaIN)~\cite{huang2017arbitrary}. 

The whitening and coloring transform used in this work has also been used in style transfer~\cite{lu2019closed,yoo2019photorealistic} and self-supervised representation learning~\cite{ermolov2021whitening}. To our knowledge, this is the first time it is being used for style injection. 

%!TEX root = main.tex
\section{Approach}

Style injection aims at modifying the statistics of the activations $\mathbf{X} \in \mathbb{R}^{C \times H W}$ at a specific layer in the generator, where $(C, H, W)$ are (number of channels, width, height) respectively. We first briefly review AdaIN style injection, then present our proposed AdaWCT generalization. 

\subsection{Adaptive Instance Normalization}
\label{sec:adain}

Adaptive Instance Normalization~\cite{huang2017arbitrary}, or AdaIN, modulates the activations $\mathbf{X}$ by 
\begin{equation}
  \label{eqn:adain}
  \mathbf{\tilde X}_\mathrm{AdaIN} = \mathrm{diag}(\boldsymbol{\sigma}) \mathrm{diag}(\nicefrac{1}{\hat{\sigma}(\mathbf{X}))}) (\mathbf{X}-\hat{\mu}(\mathbf{X})\mathbf{1}^\intercal) + \boldsymbol{\mu} \mathbf{1}^\intercal \,,
\end{equation}
where $\hat{\mu}(\cdot)$ and $\hat{\sigma}(\cdot)$ compute the row-wise mean and standard deviation vectors respectively, $\mathbf{1} \in \mathbb{R}^{HW \times 1}$ is an all-ones vector, and the $\mathrm{diag}(\mathbf{v})$ operator creates a diagonal matrix from vector $\mathbf{v}$. Here, $(\boldsymbol{\mu}, \boldsymbol{\sigma})$ are learned parameters. 

\subsection{Adaptive WCT Normalization}

Our AdaWCT approach is a generalization of AdaIN which takes into account the correlations between features maps and employs a whitening and coloring transformation (WCT) for style injection. There are two main changes over AdaIN: 1) instead of normalizing by $\mathrm{diag}(\nicefrac{1}{\hat{\sigma}(\mathbf{X}))})$ as in  eq.~\ref{eqn:adain}, we compute the whitening matrix $\mathbf{W}$; 2) instead of modulating by $\mathrm{diag}(\boldsymbol{\sigma})$, we do so with a (learned) coloring matrix $\boldsymbol{\Gamma}$. Mathematically, eq.~\ref{eqn:adain} becomes
\begin{equation}
  \label{eqn:AdaWCT} 
  \mathbf{\tilde X}_\mathrm{AdaWCT} =
  \boldsymbol{\Gamma}\mathbf{W}(\mathbf{X}-\hat{\mu}(\mathbf{X})\mathbf{1}^\intercal) + \boldsymbol{\mu} \mathbf{1}^\intercal \,.
\end{equation}

The following two subsections describe how we obtain the whitening and coloring matrices $\mathbf{W}$ and $\boldsymbol{\Gamma}$ respectively. 

\subsection{Whitening matrix}

The whitening matrix $\mathbf{W} = \mathbf{\Sigma}^{-\nicefrac{1}{2}}$ can be computed from the covariance matrix of the activations
\begin{equation}
  \label{eqn:covariance-matrix}
  \boldsymbol{\Sigma} = \frac{1}{HW-1} \bar{\mathbf{X}}\bar{\mathbf{X}}^\intercal \,,
\end{equation}
where $\bar{\mathbf{X}} = \mathbf{X}-\hat{\mu}(\mathbf{X})\mathbf{1}^\intercal$. 
Since doing so using singular value decomposition (SVD) can be unstable~\cite{Ionescu2015}, we instead follow \cite{Peihua2017TowardFasterTraining} and employ the Newton-Schulz iterative method to compute the whitening matrix 
without having to perform an eigendecomposition. The procedure is summarized in alg.~\ref{alg:newton-schulz}.

\begin{algorithm}
  \label{alg:newton-schulz}
    \SetAlgoLined
    \DontPrintSemicolon
    \KwIn{Covariance matrix $\boldsymbol{\Sigma}$, number of iterations $n$, $\epsilon > 0$}
    \KwOut{Whitening matrix estimate $\mathbf{W} \approx \mathbf{\Sigma}^{\nicefrac{1}{2}}$}
    $\mathbf{\Sigma} \leftarrow \mathbf{\Sigma} + \epsilon\mathbf{I}$ \tcp*{pre-conditioning}\label{eqn:shrink}
    $\mathbf{Y_{0}} \leftarrow\nicefrac{\boldsymbol{\mathrm{\Sigma}}}{\left \|  \boldsymbol{\mathrm{\Sigma}} \right \|_{F}}$ \tcp*{pre-compensation}\label{eqn:prenorm}
    $\boldsymbol{\mathrm{Z_0}} \leftarrow \boldsymbol{\mathrm{I}}$ \\
    \For{$i=0:n-1$}{        
      $ \mathbf{Y}_{i+1} \leftarrow \frac{1}{2}\mathbf{Y}_{i}(3\mathbf{I} - \mathbf{Z}_{i}\mathbf{Y}_i)$ 
      $ \mathbf{Z}_{i+1} \leftarrow \frac{1}{2}(3\mathbf{I} - \mathbf{Z}_{i}\mathbf{Y}_{i})\mathbf{Z}_i $ 
    }
    $\mathbf{W} \leftarrow \nicefrac{\mathbf{Z}_n}{\left \|  \mathbf{\mathrm{\Sigma}} \right \|_{F}^{\nicefrac{1}{2}}}$ \tcp*{post-compensation} \label{eqn:postcomp}
    \caption{Newton-Schulz iterative method.}
\end{algorithm}

\paragraph*{Pre-conditioning}
In order to improve the stability of the procedure, a shrinkage operator~\cite{Schafer2005Shrinkage} which improves the conditioning of $\mathbf{\Sigma}$ is first applied (l.~\ref{eqn:shrink} in alg.~\ref{alg:newton-schulz}). 

\paragraph*{Pre-compensation}
According to \cite{Higham2008}, the Newton-Shulz method locally converges only if $\left \| \mathbf{I} - \mathbf{\Sigma} \right \|_{p} < 1$ for $p=1, 2$, or $\infty$. In order to ensure this criterion is respected, the covariance matrix $\mathbf{\Sigma}$ is first normalized by its Frobenius norm~\cite{2018Decorrelated_Batch_Normalization} (l.~\ref{eqn:prenorm} in alg.~\ref{alg:newton-schulz}). 

\paragraph*{Post-compensation}
While the pre-compensation step ensures convergence, at the final step $\mathbf{Z}_n \approx \left \|\mathbf{\Sigma} \right \|_{F}^{\nicefrac{1}{2}} \mathbf{\Sigma}^{-\nicefrac{1}{2}}$ instead of the whitening matrix $\mathbf{\Sigma}^{-\nicefrac{1}{2}}$. We therefore post-compensate (l.~\ref{eqn:postcomp} of alg.~\ref{alg:newton-schulz}) by dividing with $\left \|\mathbf{\Sigma} \right \|_{F}^{\nicefrac{1}{2}}$. 

\paragraph*{Uniform normalization}
We found that the numerical compensation steps introduced by \cite{Peihua2017TowardFasterTraining} only work in cases where the matrix norm $\left \|\mathbf{\Sigma} \right \|_{F}$ is small. This is because these steps amplify any error introduced by the iterative method. We address this by pre-normalizing
$\bar{\mathbf{X}}$, dividing it with its standard deviation across all channels and pixels, just before computing the covariance matrix (eq.~\ref{eqn:covariance-matrix}). Since this rescaling is uniform, it leaves the output of alg.~\ref{alg:newton-schulz} unchanged, while consistently avoiding large matrix norms. 

\subsection{Coloring matrix}
\label{sec:coloring-matrix}

Similar to the way $(\boldsymbol{\mu}, \boldsymbol{\sigma})$ are learned parameters for AdaIN (sec.~\ref{sec:adain}), $(\boldsymbol{\mu}, \boldsymbol{\Gamma})$ are learned in our approach by the projection
\begin{equation}
\label{eqn:projection} 
\left( \boldsymbol{\mu}, \boldsymbol{\Gamma} \right) = \mathbf{\Phi}(\mathbf{w}) 
\end{equation}
of a latent vector $\mathbf{w}$ obtained from a standard mapping network~\cite{StyleGAN2018,choi2020stargan}. Here, $\boldsymbol{\Phi}$ can be an affine projection layer as in \cite{huang2017arbitrary}. While a coloring matrix should, strictly speaking, be symmetric positive semi-definite, we did not find this constraint necessary or helpful in practice. 

% PROBABLY NOT SUPER ESSENTIAL
%Fig.~\ref{fig:module_fig} illustrates how the style injection module operates inside the generator network. 
% \begin{figure}
%   \centering
%    \includegraphics[width=1.0\linewidth]{figures/AdaWCT.png}
%    \caption{Style injection using our AdaWCT module. A projection $\boldsymbol{\Phi}$ of the latent vector $w$ to a color matrix $\mathbf{\Gamma}$ is performed in order to allow the coloring process of the whitened feature maps.}
%    \label{fig:module_fig}
% \end{figure}

\subsection{Group-wise whitening and coloring}
\label{sec:groupwise-wct}

Unfortunately, generating a full coloring matrix $\mathbf{\Gamma}$ creates a parameter explosion in the generator. Take the typical example of a 256-channel feature map and 512-dimensional style vector. The total number of parameters in the projection subnetwork $\mathbf{\Phi}$ must therefore be at least $256\times256\times512 \approx 33\text{M}$. Since several style injection operations take place within a network, this quickly becomes intractable. 

We therefore employ group-wise whitening and coloring~\cite{2018Decorrelated_Batch_Normalization, OrthogonalWeightNormalization2017} to reduce the total number of parameters. The idea is to split (along the channel dimension) the activation maps $\mathbf{X} \in \mathbb{R}^{C \times HW}$ into $n$ subgroups $\mathbf{G}_j \in \mathbb{R}^{G \times HW}$, $j=\{1, ..., n\}$, that will be whitened and subsequently colored independently of each other. Here, $G = \nicefrac{C}{n}$ is the number of channels in each group.

% We denote the \emph{grouping} operator $\Omega$ as 
% %
% \begin{equation}
%   \label{eqn:grouping}
%   \Omega(\mathbf{X}) : \mathbf{X} \in \mathbb{R}^{C \times H \times W}\rightarrow \mathbf{X} \in \mathbb{R}^{n \times G \times H W} \,,
% \end{equation}
% %
% where $n$ is the number of groups, and $G$ the number of channels in each group such that $n=\nicefrac{C}{G}$. Denoting $\mathbf{G}_i \in \mathbb{R}^{G \times H W}$ with $i=\{1, ..., n\}$, group-wise correlation matrices $\mathbf{\Sigma}_i$ can thus be computed as
% %
% \begin{equation}
%   \label{eqn:sub_cor_matrices}
%   \boldsymbol{\Sigma}_{j} \in \mathbb{R}^{G \times G} = \psi(\mathbf{G}_{j})\psi(\mathbf{G}_{j})^\intercal \,,
% \end{equation} 
% % 
% where the $\psi$ operator flattens its input to a 1-D vector. 
The process is summarized in alg.~\ref{alg:group-adawct}. Here, we denote $(\boldsymbol{\mu}, \mathbf{\Gamma}_B) = \mathrm{reshape}(\mathbf{\Phi}(\mathbf{w}))$ as the group-wise projection of style vector $\mathbf{w}$, where $\mathbf{\Gamma}_B$ corresponds to the output of the projection subnetwork rearranged into a block-diagonal matrix (where each block has dimension $G \times G$). 

Note that this group-wise WCT operation discards all correlations between activations belonging to different groups. Varying the group size $G \in [1, C]$ affects the \emph{degree} of WCT applied to the features within the network. If $G=C$, full whitening and coloring is applied. Conversely, $G=1$ degrades naturally to AdaIN (sec.~\ref{sec:adain}). 

\begin{algorithm}
  \label{alg:group-adawct}
    \SetAlgoLined
    \DontPrintSemicolon
    \KwIn{Activations $\mathbf{X}$, style vector $\mathbf{w}$, $n=\frac{C}{G}$ groups }
    \KwOut{Colored activations $\mathbf{\tilde{X}}$}
    \tcp{compute group-wise correlation matrices}
    $\mathbf{G}_{j \in \{1, \dotsc, n\}} \leftarrow \mathbf{X}$\tcp*{split into groups}

    \For{$j=1:n$}{
    %   $\boldsymbol{\Sigma}_{j} \in \mathbb{R}^{G \times G} \leftarrow \psi(\mathbf{G}_{j})\psi(\mathbf{G}_{j})^\intercal$
      $\mathbf{\bar{G}}_j \leftarrow \mathbf{G}_j - \hat{\mu}(\mathbf{G}_j)\mathbf{1}^\intercal$ 
      $\boldsymbol{\Sigma}_{j} \in \mathbb{R}^{G \times G} \leftarrow (\nicefrac{1}{HW-1})\mathbf{\bar{G}}_j\mathbf{\bar{G}}_j^\intercal$ \tcp*{eq.~\ref{eqn:covariance-matrix}}
      $\mathbf{W}_{j} \leftarrow \textrm{Newton-Schulz}(\mathbf{\Sigma}_{j})$ \tcp*{alg.~\ref{alg:newton-schulz}}
    }
    % Construction de la matrice par blocs à l'aide des sous matrices $\boldsymbol{\Sigma}_{j}$ pour $j \in \{1, \dotsc, n\}$.
    $\boldsymbol{W}_B = 
    \begin{bmatrix}
    \boldsymbol{W}_{1} & & \mathbf{0} \\
    & \ddots & \\
    \mathbf{0} & & \boldsymbol{W}_{n}\\
    \end{bmatrix}
    \in \mathbb{R}^{C \times C}$

    $\mathbf{Z} \leftarrow \mathbf{W}_B(\boldsymbol{\mathrm{X}}-\hat{\mu}(\mathbf{X})\mathbf{1}^\intercal) $ \tcp*{whitening}
    
    $(\boldsymbol{\mu}, \boldsymbol{\Gamma}_B) \leftarrow \mathrm{reshape}(\mathbf{\Phi}_B(\mathbf{w}))$ \tcp*{projection}
    $\mathbf{\tilde X} \leftarrow \boldsymbol{\mathbf{\Gamma}}_B\mathbf{Z} + \boldsymbol{\mu} \mathbf{1}^\intercal$ \tcp*{coloring}

    \caption{Grouped AdaWCT style injection. }
\end{algorithm}

%!TEX root = main.tex
\section{Experiments}

We now present quantitative and qualitative experiments to validate our method.

\begin{table}[t]
    \centering
    \begin{tabular}[t]{lcccc}
      \toprule 
      & \multicolumn{2}{c}{Reference-guided} & \multicolumn{2}{c}{Latent-guided} \\
      & AdaIN & AdaWCT & AdaIN & AdaWCT  \\
      \midrule
      FID$_\downarrow$ & 19.78 & \textbf{16.20}  & 16.18 & \textbf{13.07}  \\
      LPIPS$_\uparrow$ & \textbf{0.431} & \textbf{0.434}  & 0.450 & \textbf{0.476}  \\
      \bottomrule
    \end{tabular}
    \caption{Reference- and latent-guided quantitative results. Quantitative FID$_\downarrow$ and LPIPS$_\uparrow$ metrics on the AFHQ dataset comparing our AdaWCT with AdaIN. Numbers correspond to the average obtained on all the domains for each metric.}
    \label{tab:quant}
\end{table}

\begin{figure}[t]
  \centering
   \includegraphics[width=1.0\linewidth]{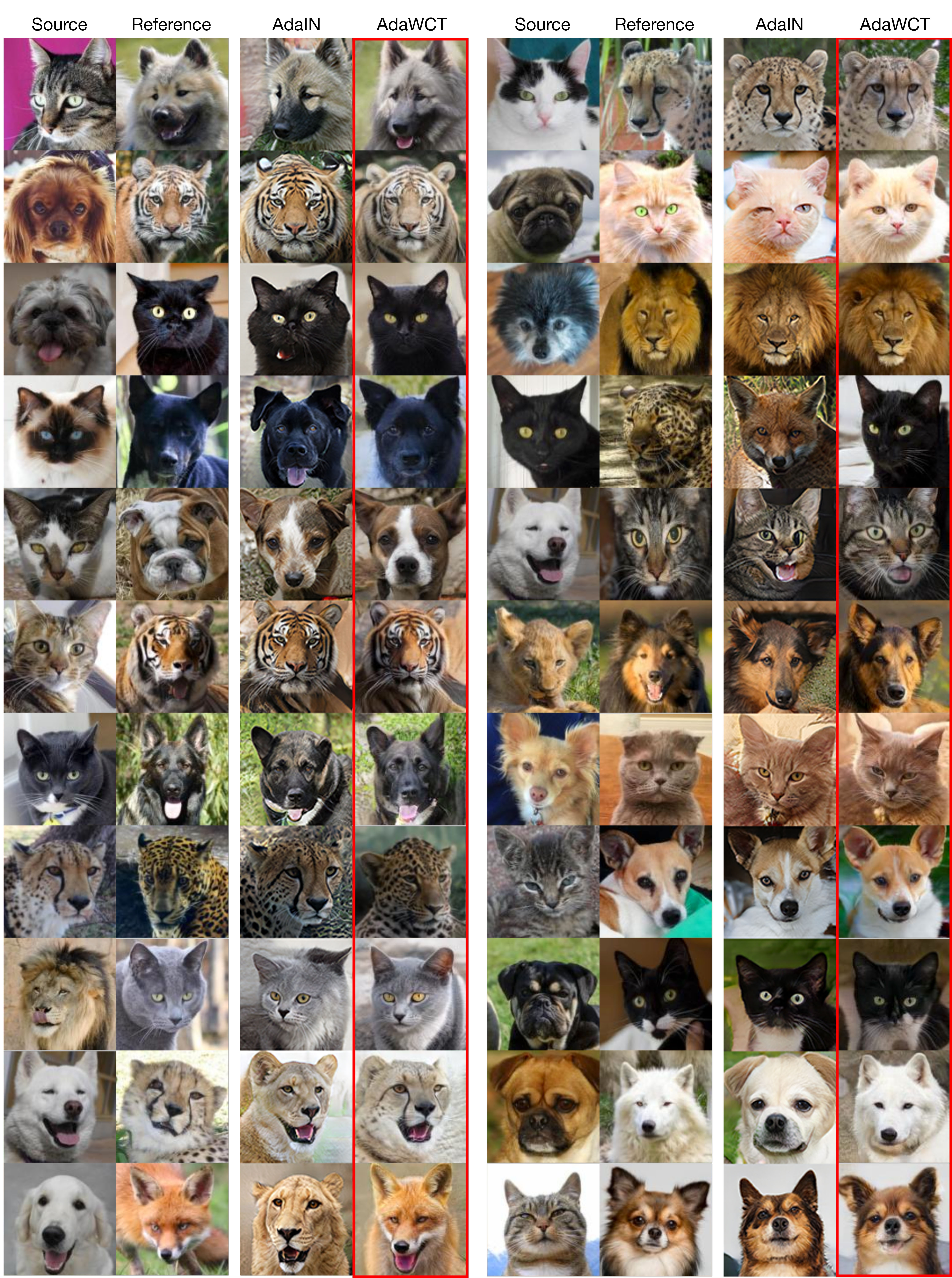}
   \caption{Reference-guided qualitative results. ``Source'' refers to the structure (pose, orientation, shape) while ``reference'' is the style (texture, color and domain) which needs to be preserved in the result.}
   \label{fig:ref-guided}
\end{figure}

\begin{figure}[t]
  \centering
   \includegraphics[width=1.0\linewidth]{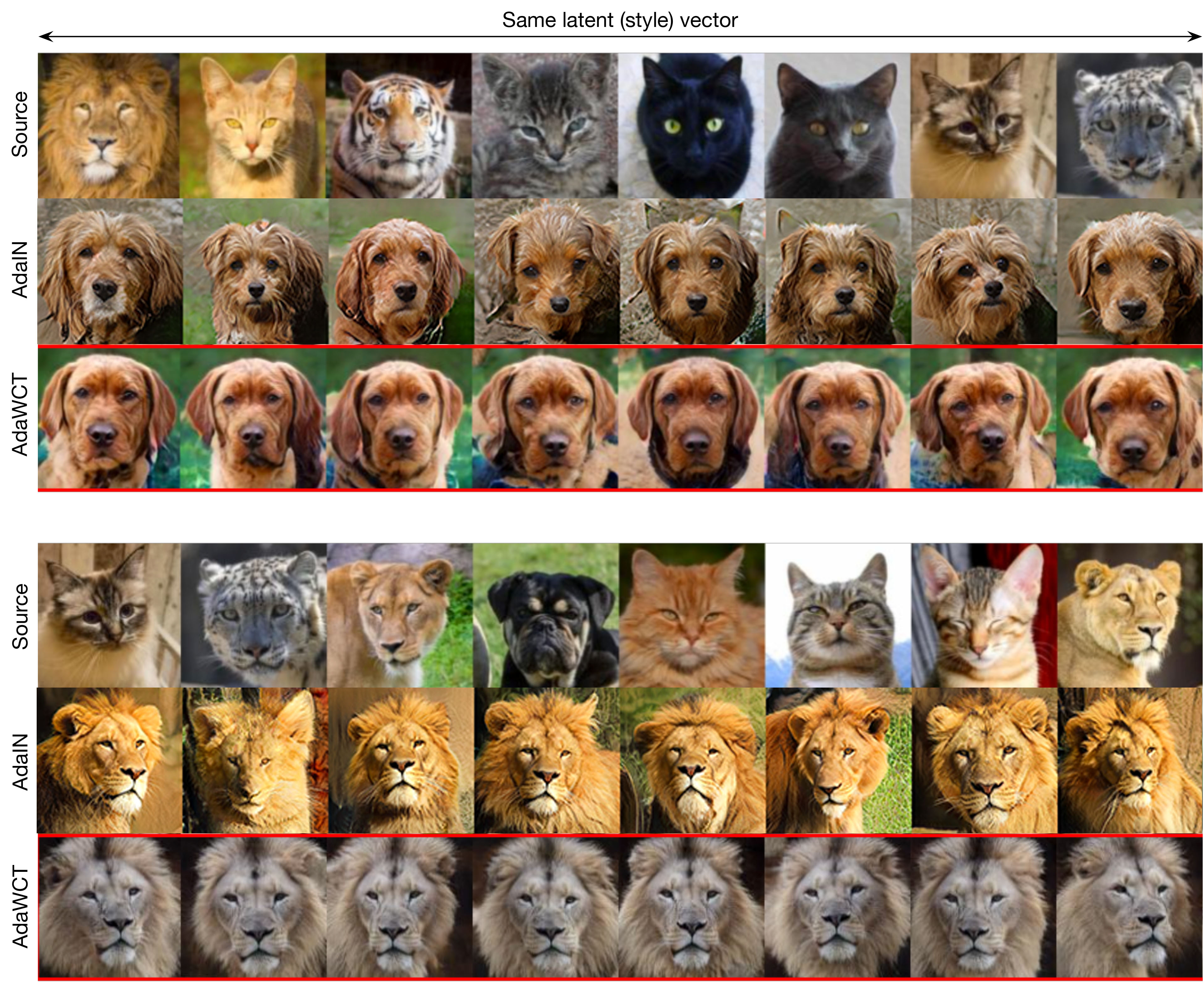}
   \caption{Latent-guided qualitative results. The same random style vector is used for each source image. First row contains the source images while second and third rows are the images generated using AdaIN and our method respectively.}
   \label{fig:latent-guided}
   \vspace{-1em}
\end{figure}

\begin{table}[t]
  \centering
    \begin{tabular}[t]{ccccc}
        \toprule
        & \multicolumn{2}{c}{Reference-guided} & \multicolumn{2}{c}{Latent-guided} \\
        $G$ & FID$_\downarrow$ & LPIPS$_\uparrow$ & FID$_\downarrow$ & LPIPS$_\uparrow$ \\
        \midrule
        1 & 24.98 & 0.309 & 18.23 & 0.342 \\
        4 & 20.79 & 0.328 & 16.65 & 0.359 \\
        8 & 20.48 & 0.338 & 20.48 & 0.368 \\
        16 & 18.22 & 0.375 & \textbf{14.02} & \textbf{0.410} \\
        32 & 17.80 & 0.370 & 14.64 & 0.399 \\
        64 & \textbf{16.63} & \textbf{0.382} & 14.24 & 0.408 \\
        \bottomrule
        \end{tabular}
    \caption[]{FID$_\downarrow$ (lower is better) and LPIPS$_\uparrow$ (higher is better) metrics as a function of group size $G$ for both reference- and latent-guided scenarios. Numbers correspond to the average obtained on all the domains for each metric. Here, results are computed on images of $128\times128$ resolution (hence results are different than in tab.~\ref{tab:quant}).}
    \label{tab:group-size}
\end{table}

\subsection{Network and dataset} 

We compare our AdaWCT approach with AdaIN within the framework of StarGANv2~\cite{choi2020stargan}, which allows for unpaired image-to-image translation across multiple domains. To adapt StarGANv2 to our AdaWCT, we simply replace all the AdaIN layers in the generator by our AdaWCT module. The other architectural components (style encoder, mapping network, discriminator) are unchanged, and so is the training regimen (loss functions, hyperparameters, etc.). 

We present results on the \enquote{Animal Faces-HQ} (AFHQ) dataset~\cite{choi2020stargan}. This dataset contains images from three domains: cats, dogs, and wild animals (mix of lion, fox, tiger, cheetah, wolf, jaguar, and leopard). Each domain is composed of 4,500/500 images for train/test respectively. As in \cite{choi2020stargan} and unless otherwise noted, all results here are reported on images of $256\times256$ resolution. 

\subsection{Quantitative and qualitative evaluation}
\label{sec:quant-qual-eval}

We evaluate the quality and diversity of results obtained with AdaWCT compared to AdaIN using the FID~\cite{FID2017} and LPIPS~\cite{BICYCLEGAN2018} metrics. Following \cite{choi2020stargan}, we show results in both 1) ``reference-guided'' (where the target style is obtained from a reference image); and 2) ``latent-guided'' (where the target style is sampled randomly) scenarios. Quantitatively, table \ref{tab:quant} presents FID and LPIPS metrics for the reference- and latent-guided scenarios. The AdaIN results are taken directly from the official StarGANv2 official pytorch implementation\footnote{\url{https://github.com/clovaai/stargan-v2}}. AdaWCT (here with $G=64$) outperforms AdaIN quantitatively all metrics except LPIPS for the reference-guided scenario. For example, in the reference-based case, AdaWCT results in an absolute FID decrease of 3.58 (-20\%). Similar improvements are observed in the latent-based case. This impact can be visualized qualitatively in figs \ref{fig:ref-guided}--\ref{fig:latent-guided} for the reference- and latent-guided scenarios respectively. In fig.~\ref{fig:ref-guided}, note how the AdaWCT results more closely match the appearance of the reference and the pose of the source, and have fewer artifacts than their AdaIN counterparts. In fig.~\ref{fig:latent-guided}, observe how the resulting images better match the pose of the source while showing fewer artifacts. 

\subsection{Sensitivity analysis}

 % mentioned that the WCT style injection was performed group-wise in order to limit the number of parameters in the generator network. 

We now study the impact of group size $G$ (sec.~\ref{sec:groupwise-wct}) on the quality of the results. Table~\ref{tab:group-size} shows the results of an experiment on images of $128 \times 128$ resolution where $G$ is progressively increased from 1 (equivalent to AdaIN, c.f. sec.~\ref{sec:groupwise-wct}) to 64. It can be seen that increasing the group from 1 to 4 makes a large difference, but as $G$ grows there are diminishing returns. The best FIDs are obtained with $G=64$ (above which training is inefficient), the value used in the experiments in sec.~\ref{sec:quant-qual-eval}.

%!TEX root = main.tex
\section{Discussion}

This paper presented a generalization of AdaIN, dubbed AdaWCT, that we use for style injection in large GANs. To do so, a group-wise whitening and coloring transformation is presented as a drop-in replacement for AdaIN. Replacing the style injection mechanism with our method within the established StarGANv2~\cite{choi2020stargan} architecture results in significantly improved performance for both reference- and latent-guided image-to-image translation results. 

\noindent \textbf{Limitations and future work} \quad 
% \paragraph*{Limitations and future work}
So far, the AdaWCT approach has been evaluated on image-to-image applications with the StarGANv2 network. It would be interesting, as future work, to study its impact in other popular GAN architectures, for example for unconditional image generation with StyleGAN3~\cite{karras2021alias}. In addition, we also replaced the affine layers with a small, 3-layer MLP for projecting the latent style vector to the coloring parameters (sec.~\ref{sec:coloring-matrix}). Simplifying this to use a single affine projection, as is commonly done~\cite{karras2021alias,choi2020stargan}, would potentially help reducing the number of additional parameters introduced by our method. 

\noindent \textbf{Broader impact} \quad 
% \paragraph*{Broader impact}
Any GAN can be employed for pernicious use. While this work was conducted with the goal of improving image generation for creative purposes, we acknowledge this paper enables the creation of more realistic images which makes them more difficult to be detected and distinguished from real photographs. We condemn misuse of these technologies, and promote solutions such as digital authentication.

\noindent \textbf{Acknowledgements} \quad This research was funded by NSERC CRDPJ 537961-18 and Gearbox Studios, with computing resources from Compute Canada and Nvidia GPU donations.

% \appendix
% \input{prenorm}

% References should be produced using the bibtex program from suitable
% BiBTeX files (here: strings, refs, manuals). The IEEEbib.bst bibliography
% style file from IEEE produces unsorted bibliography list.
% -------------------------------------------------------------------------
\bibliographystyle{IEEEbib}
\bibliography{egbib}

\begin{thebibliography}{10}

\bibitem{goodfellow2014generative}
I.~Goodfellow, J.~Pouget-Abadie, M.~Mirza, B.~Xu, D.~Warde-Farley, S.~Ozair,
  A.~Courville, and Y.~Bengio,
\newblock ``Generative adversarial nets,''
\newblock in {\em NeurIPS}, 2014.

\bibitem{Isola2017ImagetoImageTW}
P.~Isola, J.-Y. Zhu, T.~Zhou, and A.~A. Efros,
\newblock ``Image-to-image translation with conditional adversarial networks,''
\newblock in {\em CVPR}, 2017.

\bibitem{huang2017arbitrary}
X.~Huang and S.~Belongie,
\newblock ``Arbitrary style transfer in real-time with adaptive instance
  normalization,''
\newblock in {\em ICCV}, 2017.

\bibitem{choi2018stargan}
Y.~Choi, M.~Choi, M.~Kim, J.-W. Ha, S.~Kim, and J.~Choo,
\newblock ``Stargan: Unified generative adversarial networks for multi-domain
  image-to-image translation,''
\newblock in {\em CVPR}, 2018.

\bibitem{StyleGAN2018}
T.~Karras, S.~Laine, and T.~Aila,
\newblock ``A style-based generator architecture for generative adversarial
  networks,''
\newblock in {\em CVPR}, 2019.

\bibitem{choi2020stargan}
Y.~Choi, Y.~Uh, J.~Yoo, and J.-W. Ha,
\newblock ``Stargan v2: Diverse image synthesis for multiple domains,''
\newblock in {\em CVPR}, 2020.

\bibitem{goodfellow}
I.~Goodfellow, J.~Pouget-Abadie, M.~Mirza, B.~Xu, D.~Warde-Farley, S.~Ozair,
  A.~Courville, and Y.~Bengio,
\newblock ``Generative adversarial nets,''
\newblock in {\em NeurIPS}, 2014.

\bibitem{karras2020analyzing}
T.~Karras, S.~Laine, M.~Aittala, J.~Hellsten, J.~Lehtinen, and T.~Aila,
\newblock ``Analyzing and improving the image quality of stylegan,''
\newblock in {\em CVPR}, 2020.

\bibitem{karras2021alias}
T.~Karras, M.~Aittala, S.~Laine, E.~H{\"a}rk{\"o}nen, J.~Hellsten, J.~Lehtinen,
  and T.~Aila,
\newblock ``Alias-free generative adversarial networks,''
\newblock {\em NeurIPS}, 2021.

\bibitem{MUNIT}
X.~Huang, M.-Y. Liu, S.~Belongie, and J.~Kautz,
\newblock ``Multimodal unsupervised image-to-image translation,''
\newblock in {\em ECCV}, 2018.

\bibitem{pix2pix}
P.~Isola, J.-Y. Zhu, T.~Zhou, and A.~A. Efros,
\newblock ``Image-to-image translation with conditional adversarial networks,''
\newblock in {\em CVPR}, 2017.

\bibitem{DRIT}
H.-Y. Lee, H.-Y. Tseng, J.-B. Huang, M.~Singh, and M.-H. Yang,
\newblock ``Diverse image-to-image translation via disentangled
  representations,''
\newblock in {\em ECCV}, 2018.

\bibitem{msgan}
Q.~Mao, H.-Y. Lee, H.-Y. Tseng, S.~Ma, and M.-H. Yang,
\newblock ``Mode seeking generative adversarial networks for diverse image
  synthesis,''
\newblock in {\em CVPR}, 2019.

\bibitem{cycleGan}
J.-Y. Zhu, T.~Park, P.~Isola, and A.~A. Efros,
\newblock ``Unpaired image-to-image translation using cycle-consistent
  adversarial networks,''
\newblock in {\em CVPR}, 2017.

\bibitem{Wgan}
M.~Arjovsky, S.~Chintala, and L.~Bottou,
\newblock ``{W}asserstein generative adversarial networks,''
\newblock in {\em ICML}, 2017.

\bibitem{wgangp}
I.~Gulrajani, F.~Ahmed, M.~Arjovsky, V.~Dumoulin, and A.~C. Courville,
\newblock ``Improved training of wasserstein gans,''
\newblock in {\em NeurIPS}, 2017.

\bibitem{lsgan}
M.~Xudong, L.~Qing, X.~Haoran, L.~Raymond Y.~K., and W.~Zhen,
\newblock ``Least squares generative adversarial networks,''
\newblock in {\em ICCV}, 2017.

\bibitem{diverseGan}
D.~Yang, S.~Hong, Y.~Jang, T.~Zhao, and H.~Lee,
\newblock ``Diversity-sensitive conditional generative adversarial networks,''
\newblock in {\em ICLR}, 2019.

\bibitem{ProGanCelebA}
T.~Karras, T.~Aila, S.~Laine, and J.~Lehtinen,
\newblock ``Progressive growing of {GAN}s for improved quality, stability, and
  variation,''
\newblock in {\em CVPR}, 2018.

\bibitem{CItoI}
J.~Lin, Y.~Xia, T.~Qin, Z.~Chen, and T.-Y. Liu,
\newblock ``Conditional image-to-image translation,''
\newblock in {\em CVPR}, 2018.

\bibitem{lu2019closed}
M.~Lu, H.~Zhao, A.~Yao, Y.~Chen, F.~Xu, and L.~Zhang,
\newblock ``A closed-form solution to universal style transfer,''
\newblock in {\em ICCV}, 2019.

\bibitem{yoo2019photorealistic}
J.~Yoo, Y.~Uh, S.~Chun, B.~Kang, and J.-W. Ha,
\newblock ``Photorealistic style transfer via wavelet transforms,''
\newblock in {\em ICCV}, 2019.

\bibitem{ermolov2021whitening}
A.~Ermolov, A.~Siarohin, E.~Sangineto, and N.~Sebe,
\newblock ``Whitening for self-supervised representation learning,''
\newblock in {\em ICML}, 2021.

\bibitem{Ionescu2015}
C.~{Ionescu}, O.~{Vantzos}, and C.~{Sminchisescu},
\newblock ``Matrix backpropagation for deep networks with structured layers,''
\newblock {\em CoRR}, 2015.

\bibitem{Peihua2017TowardFasterTraining}
P.~Li, J.~Xie, Q.~Wang, and Z.~Gao,
\newblock ``Towards faster training of global covariance pooling networks by
  iterative matrix square root normalization,''
\newblock in {\em CVPR}, 2018.

\bibitem{Schafer2005Shrinkage}
J.~Sch\"{a}fer and K.~Strimmer,
\newblock ``A shrinkage approach to large-scale covariance matrix estimation
  and implications for functional genomics,''
\newblock {\em SAGMB}, 2005.

\bibitem{Higham2008}
N.~J. Higham,
\newblock {\em Functions of Matrices: {Theory} and Computation},
\newblock Soc. for Ind. and App. Math., 2008.

\bibitem{2018Decorrelated_Batch_Normalization}
L.~Huang, D.~Yang, B.~Lang, and J.~Deng,
\newblock ``Decorrelated batch normalization,''
\newblock in {\em CVPR}, 2018.

\bibitem{OrthogonalWeightNormalization2017}
L.~Huang, X.~Liu, B.~Lang, A.~W. Yu, and B.~Li,
\newblock ``Orthogonal weight normalization: Solution to optimization over
  multiple dependent stiefel manifolds in deep neural networks,''
\newblock in {\em AAAI}, 2018.

\bibitem{FID2017}
M.~Heusel, H.~Ramsauer, T.~Unterthiner, B.~Nessler, G.~Klambauer, and
  S.~Hochreiter,
\newblock ``Gans trained by a two time-scale update rule converge to a nash
  equilibrium,''
\newblock {\em CoRR}, vol. abs/1706.08500, 2017.

\bibitem{BICYCLEGAN2018}
J.~Zhu, R.~Zhang, D.~Pathak, T.~Darrell, A.~A. Efros, O.~Wang, and
  E.~Shechtman,
\newblock ``Toward multimodal image-to-image translation,''
\newblock {\em CoRR}, vol. abs/1711.11586, 2017.

\end{thebibliography}

\end{document}